\documentclass[letterpaper, 10 pt, journal, twoside]{IEEEtran}

\usepackage{cite}
\usepackage{amsmath,amssymb,amsfonts}
\usepackage{algorithmic}
\usepackage{graphicx}
\usepackage{textcomp}
\usepackage{xcolor}
\graphicspath{{./images/}}
\def\BibTeX{{\rm B\kern-.05em{\sc i\kern-.025em b}\kern-.08em
    T\kern-.1667em\lower.7ex\hbox{E}\kern-.125emX}}
\usepackage{float}

\begin{document}
\title{Snapp: An Agile Robotic Fish with 3-D Maneuverability for Open Water Swim}

\author{Timothy J.K. Ng$^{1}$, Nan Chen$^{1}$, and Fu Zhang$^{1}$ 
\thanks{Manuscript received: March 28, 2023; Revised June 14, 2023; Accepted July 18, 2023. This paper was recommended for publication by Editor Xinyu Liu upon evaluation of the Associate Editor and Reviewers' comments. This work was supported by the Hong Kong Research Grant Council (RGC) under the Early Career Scheme (ECS), project no. 27202219. \it{(Corresponding author: Fu Zhang.)}}
\thanks{The authors are with Mechatronics and Robotic Systems (MaRS) Laboratory, Department of Mechanical Engineering, The University of Hong Kong (e-mail: timng@connect.hku.hk; cnchen@connect.hku.hk; fuzhang@hku.hk
}
\thanks{© 2023 IEEE.  
Personal use of this material is permitted.  
Permission from IEEE must be obtained for all other uses, in any current or future media, including reprinting/republishing this material for advertising or promotional purposes, creating new collective works, for resale or redistribution to servers or lists, or reuse of any copyrighted component of this work in other works. Digital Object Identifier (DOI): 10.1109/LRA.2023.3308015}
}
\markboth{IEEE Robotics and Automation Letters. Preprint Version. Accepted July, 2023 }{Ng \MakeLowercase{\textit{et al.}}: Snapp: An Agile Robotic Fish with 3-D Maneuverability for Open Water Swim
}

\maketitle
\begin{abstract}
Fish exhibit impressive locomotive performance and agility in complex underwater environments, using their undulating tails and pectoral fins for propulsion and maneuverability.
Replicating these abilities in robotic fish is challenging; existing designs focus on either fast swimming or directional control at limited speeds, mainly within a confined environment. 
To address these limitations, we designed Snapp, an integrated robotic fish capable of swimming in open water with high speeds and full 3-dimensional maneuverability.   
A novel cyclic-differential method is layered on the mechanism. It integrates propulsion and yaw-steering for fast course corrections.
Two independent pectoral fins provide pitch and roll control.  
We evaluated Snapp in open water environments. We demonstrated significant improvements in speed and maneuverability, achieving swimming speeds of 1.5 m/s (1.7 Body Lengths per second) and performing complex maneuvers, such as a figure-8 and S-shape trajectory.
Instantaneous yaw changes of 15$^{\circ}$ in 0.4 s, a minimum turn radius of 0.85 m, and maximum pitch and roll rates of 3.5 rad/s and 1 rad/s, respectively, were recorded. 
Our results suggest that Snapp's swimming capabilities have excellent practical prospects for open seas and contribute significantly to developing agile robotic fishes. 
The accompanying video can be found at this link: 
https://youtu.be/1bGmlN0Jriw
\end{abstract}

\setlength{\textfloatsep}{6pt}
\setlength{\floatsep}{8pt}
\section{Introduction}
Fish display exceptional agility in water with their fast swimming and remarkable maneuverability\cite{LauderSummary}.
Their undulatory swimming has attracted researchers' interest with the promise of an underwater fish robot with better swimming power, energetic efficiency, and control authority. 
For example, Tuna fish have cruising speeds of 1.7 m/s and can sustain long distance swimming \cite{swimspeedtuna}. 
The growing need for a controllable, field-deployable, fish-inspired robotic platform for underwater applicatons\cite{Locomotion} has encouraged many robot fish designs. To succeed in open water swims, both speed and maneuverability are important to design an agile fish robot. 
This agility depends on their structures and can be categorized into two groups: highly articulated and simply articulated.   
\begin{figure}[t]
    \centering
    \includegraphics[width=\linewidth, height = 2in]{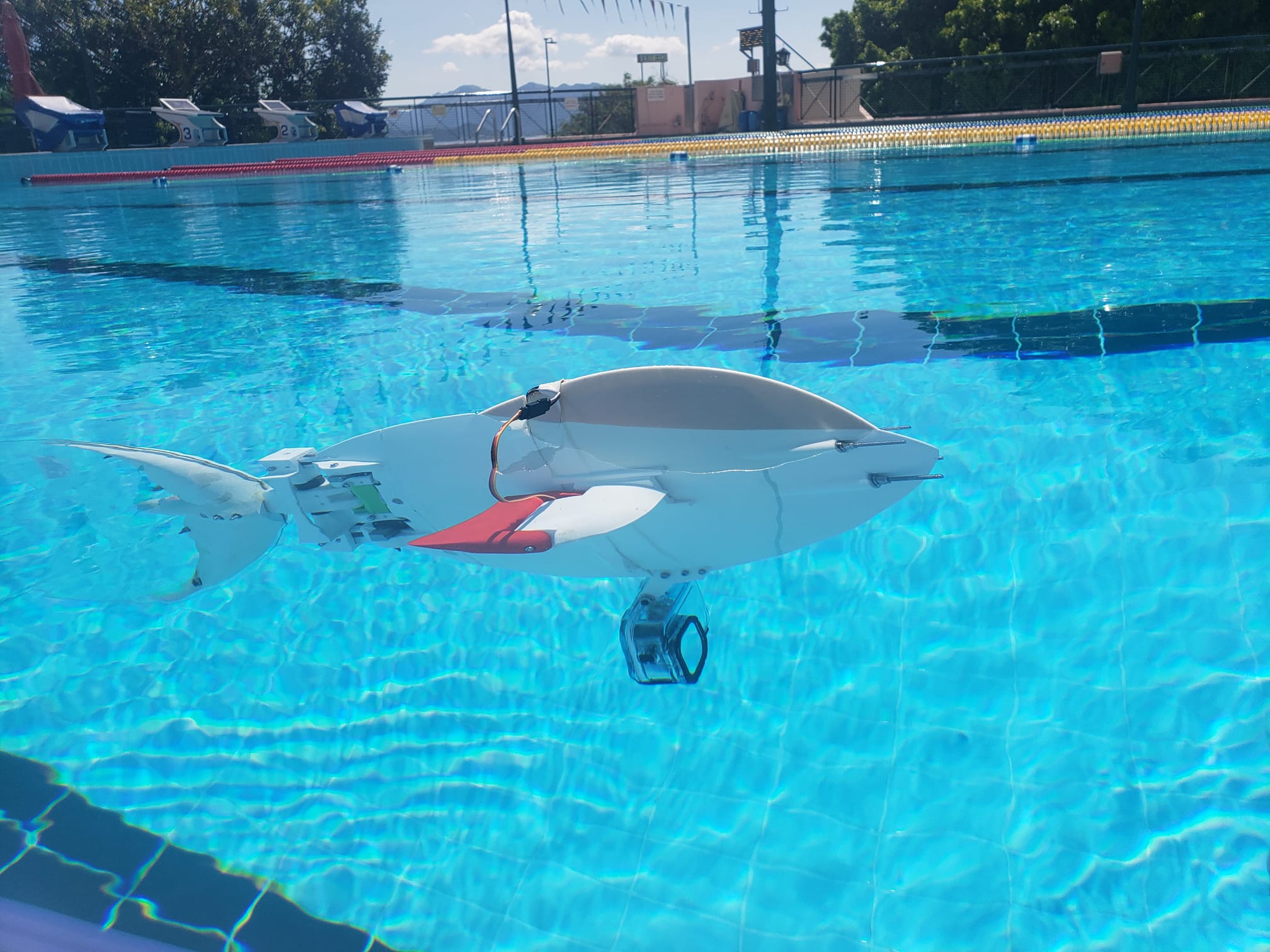}
    \caption{Snapp, the agile robotic fish deployed in an open water environment with an underwater camera attached to the bottom. It is statically stable in an upright form and can carry payloads of up to 1kg.} 
    \label{fig: Snapshot}
\end{figure}

\begin{figure}[t]
    \centering
    \includegraphics[height = 1.5in]{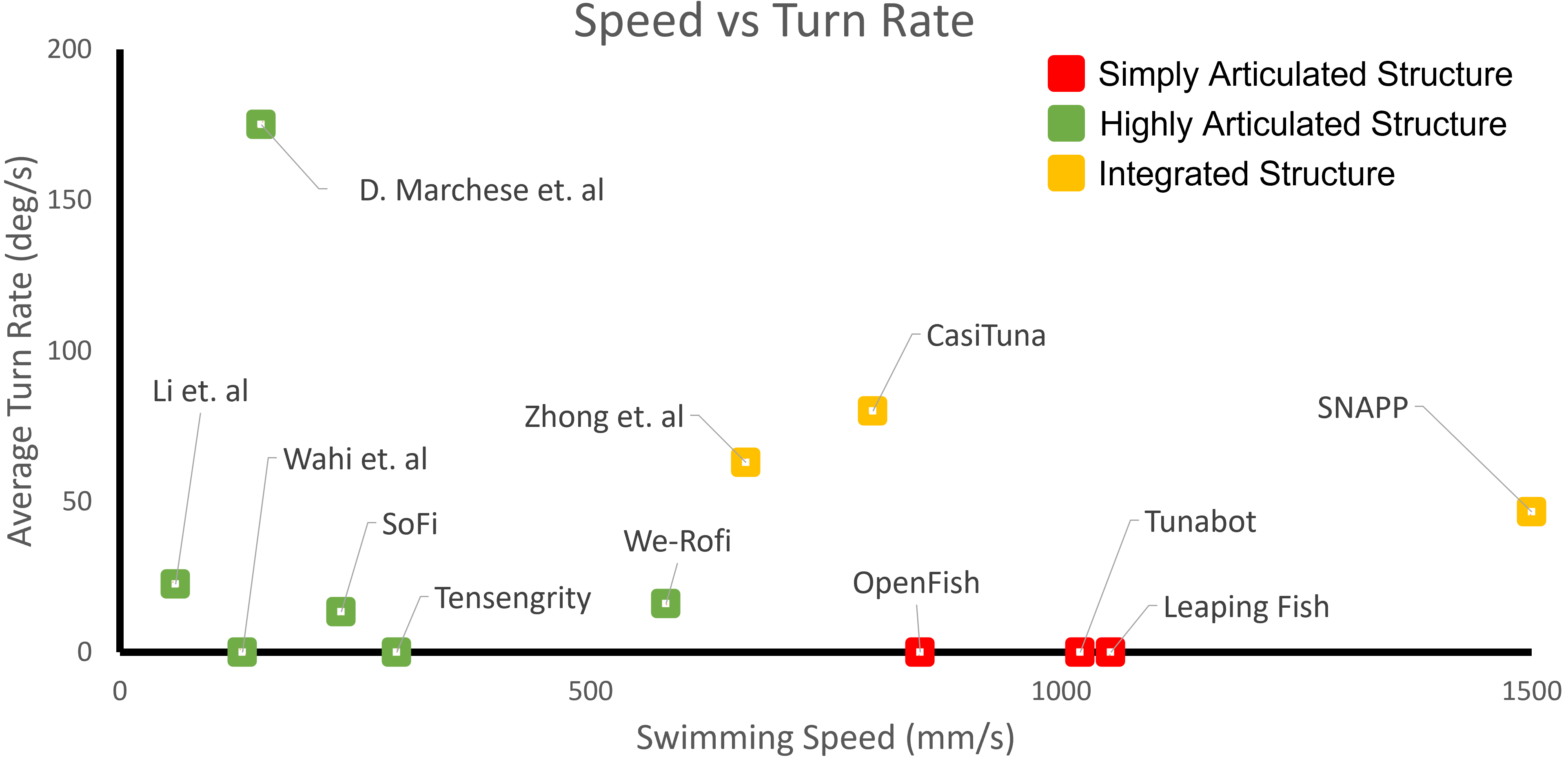}
    \caption{Distribution of robotic fish emphasizing either fast turns or pure speeds without turns. \cite{CasiTuna,LeapingFish,weRofi,tunabot,Sofi,SoroFastEscape,fishturn_IJAR,cuhk,OpenFish}}
    \label{Distribution of Robot fish} 
\end{figure}

Highly articulated structures have a large degree of control over the shape of the robot fish, mimicking waveforms of natural fishes, specifically carangiform motion \cite{weRofi,TensengrityFish, fishturn_IJAR, Sofi}
These mechanisms include parallel-series linkages,  wire-driven tendon-based skeletons, or soft visco-elastic designs. 
They have great control over the fish's morphology, with C-shapes yielding fast turns and decent swimming performance with S-shapes. 
However, they exhibit poor power transmission and a reduced actuation frequency ceiling resulting in lower swimming speeds. For wire-driven joint networks, the model complexity and slack complicate the control of the robot.
In contrast, simply-articulated structures focus on one actuators driving a scotch-yoke mechanism with high-stiffness fins to recreate fish-like swimming. \cite{LeapingFish,tunabot,OpenFish}.
This mechanism structure effectively replicates the high frequency tail-beat of Tuna, leading to fast swimming performance.
However, this mechanism maneuvers poorly because of its limited control over its morphology. The high-frequency oscillations also cause hydrodynamic recoil, resulting in lateral instability. \cite{recoil,swim_stability}. As the mechanism structure is closely related to its swimming abilities, it is important to integrate both maneuverability and speed in the design process. Figure \ref{Distribution of Robot fish} summarizes the trade-off between speed and maneuverability of these structures. 

Some robotic fish shows potential for an agile open water swimmer. 
% SOFI \cite{Sofi} was the first open water swimmer which achieved full 3-D maneuverability and propulsion. 
\cite{CasiTuna, cuhk} have attempted integration between swimming and maneuverability. 
They build on the highly articulated structures, improving their performance with swimming speeds and turn rates of about 0.8 m/s, and 80 $^{\circ}$/s respectively. Despite the highly articulated structure, they report facing problems with the lateral stability as they reached higher speeds. 
\cite{MagnetFish,Sofi} highlight the engineering challenges for open water swimming, such as buoyancy, weight, waterproofing, and efficiency. 
Hence, it is necessary to integrate these advantages and overcome their shortcomings to create an agile fish with open-water swimming abilities.

We present Snapp, the first controllable robotic fish capable of fast swims and agile maneuvers in an open water environment. Our robot fish adopts the simply-articulated mechanism structure and imposes a novel cyclic differential control for yaw control. Attached to the mechanism is a flexible caudal fin which significantly improves maneuverability of the existing simply articulated structure. 
Compared to the state-of-the-art, Snapp has the fastest a absolute swimming speed at 1.5 m/s, and comparable maneuverability at 46.7 $^{\circ}$/s.
The pectoral fins leverages the high swimming speeds for stability and control in its pitch and roll; They solve the problem of lateral stability and provide altitude control.
As speed is essential for our system, our mechanism can tune the dimensionless Strouhal number by adjusting the frequency and amplitude parameters on-the-fly. This allows users to optimize over different tail-fin parameters quickly.
Snapp is compact and untethered, weighs 5 kg with active and passive stability mechanisms, has a 30-minute continuous operating time, and can carry a payload of up to 1kg.

Snapp's design is inspired by key carangiform and thunniform swimming features \cite{lighthill_1970, Virtual_mass, LauderSummary}: (i) lunate fin shape, (ii) rapid tail oscillations focused towards the caudal fin \cite{tunabot}, and (iii) winged-shaped pectoral fins for stability.  
Snapp's performance resembles the agility of Tuna with a swimming speed of 1.5 m/s and full 3D maneuverability. It is the first bio-inspired robotic fish to demonstrate advanced swimming capabilities in open water environments by integrating speed, maneuverability, and agility. 

\section{Methods, Materials, and Design}
\subsection{Design and Fabrication}
\begin{figure}[t]
    \centering 
    \includegraphics[ height = 2in]{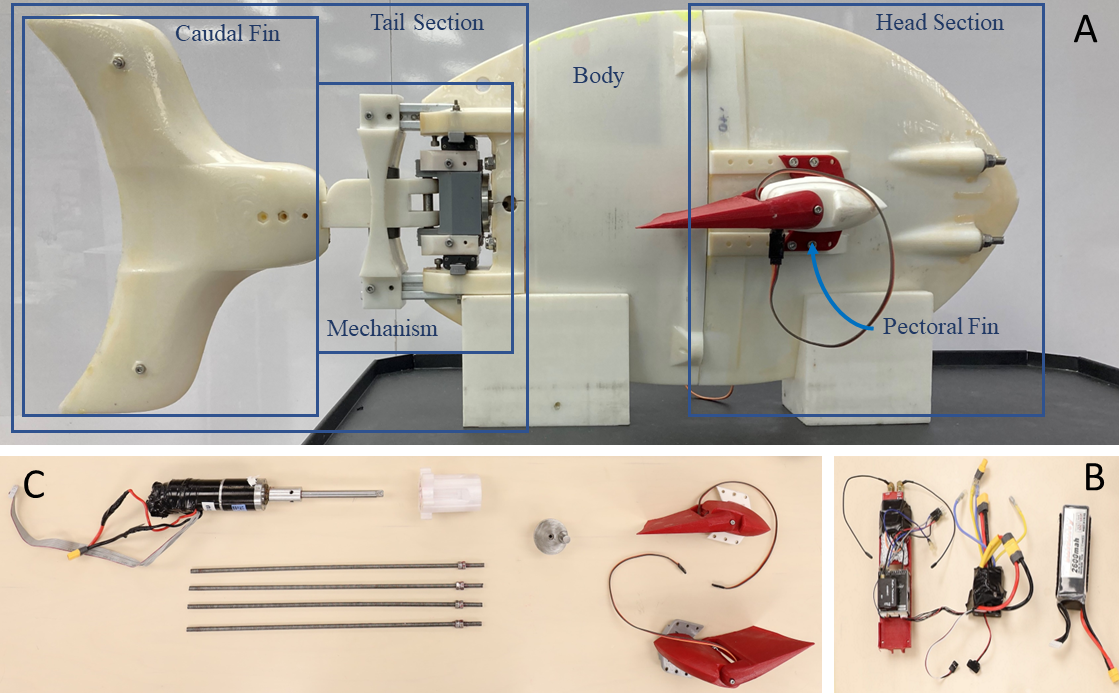}
    \caption{Detailed components of the robot fish. (A) The chassis consists of 3 major components, the (i) headpiece, which houses the electronics, the (ii) body, which the motor is mounted to and acts as a support structure; and (iii) the tail section, consisting of a scotch-yoke mechanism, pivot lever and a flexible tail. These are combined with stainless-steel tension rods to provide structural rigidity. A modular caudal tail-fin is then attached to the end of the mechanism, with the flexible portion attached directly to the caudal fin. (B) The actuators and tension rods. (C) Electrical and Electronics of Snapp. The robot measures a length of 550 mm without the tail-fin, a width of 74 mm, and a height of 292 mm, weighs 5 kg and is neutrally buoyant and swim for 30 minutes. The caudal fin has a length of 300mm, with a total length of 850 mm. }
    \label{fig:full assembly}
\end{figure}

 The fish chassis and caudal fin are made from Tough Polylactic Acid (TPLA) by 3D printing. These are coated with XTC-4D epoxy resin for waterproofing. 
 The caudal fin has a flexible component made from a 2mm Polycarbonate sheet. Snapp is powered by a Maxon 150-W, RE40 DC motor (7800 RPM, Maxon GP42C with a gear ratio of 12:1) and is controlled by a Tritonic ESC 4X motor driver via Pulse Width Modulation (PWM) and is powered by a 24-V LiPo battery. 
 The motor is mounted on a PETG piece for thermal resistance to the motor heat. Fig. \ref{fig:full assembly} details the full assembly of Snapp.
 
 Two Savox 1211-SG waterproof servos power the pectoral fins. Motor feedback is given by the HEDL 5140 incremental encoder with 500 counts per turn. The Pixhawk microcontroller processes the sensor signal with a 400Hz sampling rate. It receives user commands for the throttle, roll, pitch, and yaw and translates them into commands for the pectoral fins and the caudal fin. Underwater communication was achieved using EzUHF 433-MHz, 500-mW Transmitter, and Diversity 8-channel receiver.
\subsection{Bioinspiration and Carangiform Swimming}
We take inspiration from the carangiform mode of swimming, with the lateral motion increasing from head to tail. 
This motion $y(x,t)$ closely resembles that of a traveling wave, with $x$ being the normalized length measured from the stationary point on the body \cite{DragReduction},
\begin{equation}
    \begin{aligned}
           y(x,t)   = a(x) \sin{(kx-\omega_{m} t)},\ a(x)=c_1x + c_2 x^2
        \label{quadratic wave} 
    \end{aligned}
\end{equation}
where $k = 2\pi/\lambda$ is the wavenumber and $\lambda$, the wavelength the travelling wave, $\omega$, the oscillation frequency associated with the flapping frequency $\omega_{m}$, and $a(x)$, the amplitude envelope,
where $c_1$ and $c_2$ are adjustable parameters. 
We mimic these ratios closely from \cite{tunabot} and realize this with two rigid links consisting of the body (550mm) and the tail (150mm), combined with the flexible fin (150mm).

The lunate-shape fin of real fish \cite{lighthill_1970} inspires the caudal fin design, albeit enlarged compared to natural swimmers. The increased size increases the volume of water pushed, providing a larger lift surface for improved propulsion.
At low frequencies, it acts as a large rudder as the bending is dependent on the frequency. Fig. \ref{fig:actuation and phases}(B) and equation (\ref{equation: tail acceleration}) illustrates this effect. Though a large surface would increase drag, the increased bending at high frequencies minimizes this drag area. 
Since tuna use their pectoral fins to steer and stabilize themselves during high-speed maneuvers, we applied the same principles to the pectoral fins for Snapp.

\subsection{Buoyancy and Passive Vertical Stability}
We chose a passive method to maintain stability for energy efficiency.
A heavy bottom ensures that the fish is statically stable with slotted weights. 
These calibration weights are added until Snapp has slightly positive buoyancy. 
The large surface areas of the pectoral fins provide drag resistance to the rolling.
This combination ensures a stable system with a stable equilibrium at the vertical. 
Fig. \ref{fig: Snapshot} showcases Snapp's vertical stability. 
\subsection{Tail Mechanism Design}
\begin{figure}[t]
    \centering
    \includegraphics[ width = \linewidth]{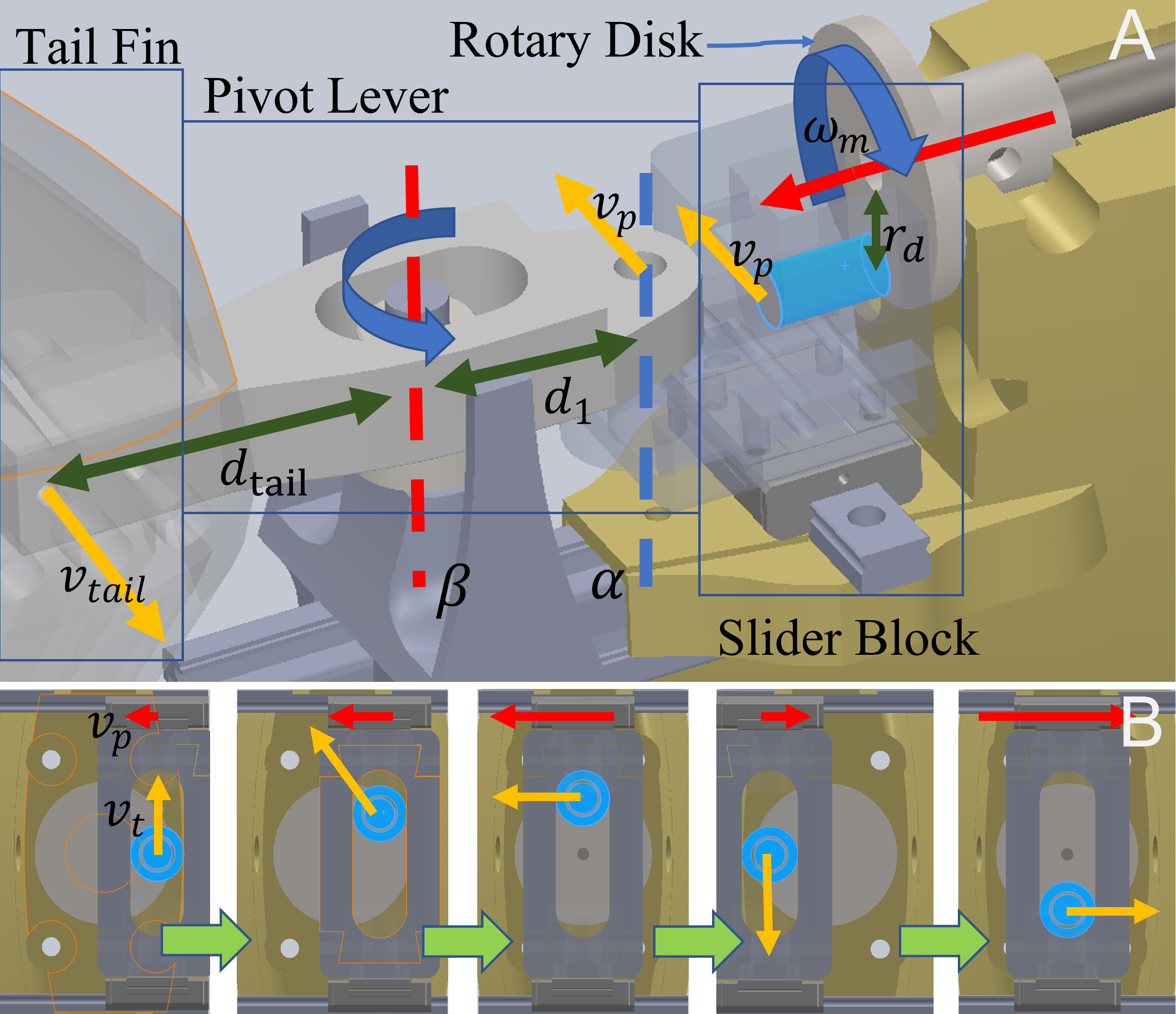}
    \caption{Tail mechanism in detail. (A)  The disk rotates in a circular motion and is connected to the slider-block slot via an extended shaft highlighted in blue. At every rotation point, the shaft on the disk is in contact with the slider block, which is constrained to move in the lateral direction via the linear rail bearings. As the motor rotates with motor speed, $\omega_{m}$, the slider moves laterally with velocity, $v_p$. The pivot lever is mounted via a shaft to the slider block at $\alpha$, which undergoes lateral translation and rotation. The $\beta$ axis is fixed to the body. The pivot lever rotates about the $\beta$ axis and can slide freely, i.e., along $d_{\rm{1}}$ causing the distance between axis $\alpha$ and $\beta$ changes in one tail beat. This results in the end velocity of the tail, $v_{tail}$. Combined, these act as a see-saw-like lever with $\beta$ as the fixed rotation axis and $V_{tail}$ as the output.
    (B) Scotch-yoke in action: Rotational to lateral motion with the tangential velocity, $V_{p}$}
    \label{fig:scotchyoke_detailed}
\end{figure}

We designed a single motor-actuated lightweight mechanism capable of reaching high tail-beat frequencies to achieve high-speed swimming. \cite{worldrecordfish} reports that speed is proportional to frequency for large caudal fins. The tail mechanism is divided into three sections, the scotch-yoke mechanism, a sliding pivot lever, and a flexible caudal fin. Fig. \ref{fig:scotchyoke_detailed} depicts the details of the tail mechanism. 
\begin{figure}[t]
    \centering
    \includegraphics[ width = \linewidth]{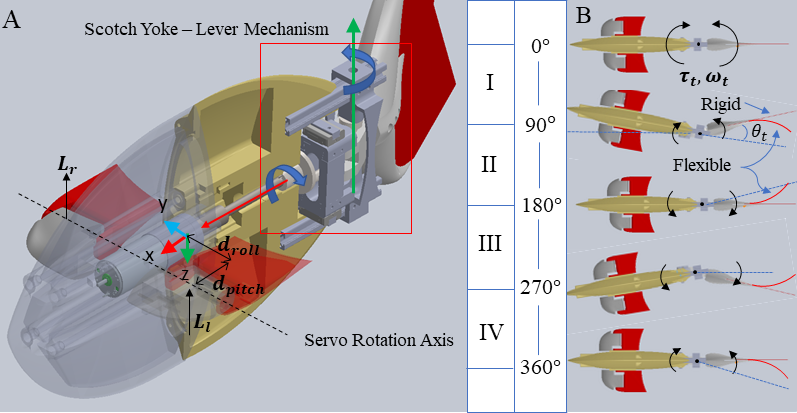}
    \caption{Actuation of the fish. (A) The pectoral fins are responsible for the dive (i.e., pitch) and roll motions and rotate about the y-axis (blue). The asymmetric angles of the fins create a roll, and similar angles create pitching moments. The motor is positioned in Snapp's head and transmits forces to the mechanism. (B) Mechanism motion and its corresponding tail motion: Phase I starts at $0^\circ$ to $90^\circ$ of the rotary disk and rotates counter-clockwise(+). The tail-fin swings right (II, $90^\circ$to $180^\circ$), back to neutral (III), swings left (IV), and back to neutral. The tail flexes to create an S-shape for fast swimming.}
    \label{fig:actuation and phases}
\end{figure}
A key parameter in the swimming speed of an undulatory fish is its Strouhal number (St) which is
\begin{equation}
    St = {fAU^{-1}}
\end{equation}
where $f$ is the tail-beat frequency governed by the motor speed, $\omega_{m}$, $A$ is the  peak-to-peak tail amplitude, and $U$ is the free flow swimming velocity. Fast swimming fish optimize swim at a $St$ number between 0.3 to 0.4\cite{Strouhal}.
We identify an optimal pair of $f$ and $A$ through iterative experiments. We found the highest speed was achieved with a frequency of 4Hz and a St number of 0.3.

The amplitude parameters, $A$ and $\theta_{\rm{t}}$ seen in Fig. \ref{fig:actuation and phases}(B) can be adjusted by changing $d_{\rm{1}}$ (Fig. \ref{fig:scotchyoke_detailed}) with the following:
\begin{equation}
    \begin{aligned}
     A =  d_{\rm{tail}} r_{\rm{d}}  d^{-1}_{\rm{1}} ,  \
    \theta_{\rm{t}} = \arctan{\left(r_{\rm{d}} d^{-1}_{\rm{1}}\right)}
    \end{aligned}
    \label{eq:angle_theta_t}
\end{equation}
where $d_{\rm{tail}}$ is the tail length from the $\beta$ axis, $r_{\rm{d}}$ the rotary disk radius, and $d_{\rm{1}}$ the distance between $\beta$ and $\alpha$ axis. The maximum tail angle of attack is 30$^{\circ}$. The flexibility of this design allows us to tune $St$ accordingly. 
The lateral undulation of the swimming fish is realized with the scotch yoke mechanism. Fig. \ref{fig:scotchyoke_detailed}(A) depicts the kinematics of the mechanism. With a fixed motor speed, the slider block moves with a sinusoidal lateral velocity, $v_{\rm{p}}$, which acts on the pivot lever. The relations are given as
\begin{equation}
    \begin{aligned}
       v_{\rm{tail}} = v_{\rm{p}} {d_{\rm{tail}}} d_{1}^{-1},\ v_{\rm{p}} = r_{\rm{d}} \omega_{\rm{m}} \cos{\theta_{\rm{m}}}
    \end{aligned}
    \label{equation: v_tail}
\end{equation}
 where $\theta_{\rm{m}}$ is the angle of the rotary disk, and $\omega_{\rm{m}}$, the motor speed. Solving for $v_{\rm{tail}}$ yields
\begin{equation}
    \begin{aligned}
 v_{\rm{tail}} = K \omega_{\rm{m}} \cos{\theta_{\rm{m}}} ,\
 K = r_{\rm{d}} {d_{\rm{tail}}} {d^{-1}_{\rm{1}}}
    \label{equation: v_tail_full}
    \end{aligned}
\end{equation}
with the mechanical constant, $K$.

Another factor for fast swimming is the S-like bends of a fish as described in Fig. \ref{fig:actuation and phases}B. We replicated this with a compliant tail of a suitable stiffness. The deflection of the flexible portion of the caudal fin depends on the caudal fin's acceleration, $a_{\rm{tail}}$. We can find $a_{\rm{tail}}$ by differentiating $v_{\rm{tail}}$ (\ref{equation: v_tail_full}) with respect to time to yield
\begin{equation}
    \label{equation: tail acceleration full}
    a_{\rm{tail}} = K (\Dot{\omega}_{\rm{m}} \cos{\theta_{\rm{m}}} + \omega_{\rm{m}}^2 \sin{\theta_{\rm{m}}} )
\end{equation}
We assume a steady-state operation during swimming. Then $\Dot{\omega}_{\rm{m}}=0$  to give $a_{\rm{tail}}$
\begin{equation}
    a_{\rm{tail}} = K \omega_{\rm{m}}^{2} \sin{\theta_{\rm{m}}}
    \label{equation: tail acceleration}
\end{equation}

A higher tail acceleration leads to larger deflections of the flexible fin, which results in an increased thrust force due to the favorable angle of attack. These features combined make Snapp an agile swimming fish. 

\section{Actuation and Control} 
The hydrodynamics of fish swimming is complex as it involves elements of vorticity and non-linear fluid dynamics. We proposed a set of control actions with outputs that are strictly increasing to stabilize and control Snapp. Fig. \ref{fig:actuation and phases}A shows the actuation mechanisms of Snapp, where the pectoral fins control roll and pitch, and the caudal fin controls yaw. We have an element-normalized control input vector
\begin{equation}
    u = [ u_{a} \;u_{s}\; u_{r}\; u_{p} ]^{T}
    \label{controlinputs}
\end{equation}
where $u_a$, $u_s$, $u_r$, and $u_p$ are the normalized motor average voltage, the normalized amplitude of the square-wave voltage for yaw control, and the normalized control inputs for roll and pitch, respectively.

\subsection{Actuation of Pitch, Roll, and Altitude}
The pectoral fins act as the control surface for pitch and roll by inducing torques about their respective axis. They are speed dependent, and are governed by the lift principle,
\begin{equation}
    \label{lift mechanics}
    L(\theta,V) = \frac{1}{2}   \rho_{\rm{w}} c_{L}({\theta}) S V^2 
\end{equation}
where the hydrodynamic parameters $c_{L}$, $V$, $S$, $\rho_{\rm{w}}$ as the lift coefficient, swimming velocity relative to the water, control surface fin area, and water density, respectively. $\theta$ is the angle of attack of the respective control surfaces, controlled by the user, and is related to the control input by 
\begin{equation}
   \begin{bmatrix}
    \theta_{l} \\ \theta_{r}
   \end{bmatrix} = K
    \begin{bmatrix}
    1& 1\\1&-1
   \end{bmatrix} 
      \begin{bmatrix}
    u_{r} \\ u_{p}
   \end{bmatrix}\
\end{equation}
where $\theta_{l}$ and $\theta_{r}$ are the angle of attack of the left and right pectoral fins, and $K$ is the constant mapping coefficient. Fig. \ref{fig:actuation and phases}(A) shows the corresponding forces and their moment arms. The fish lift ($F_{z}$), the roll ($M_{\rm{roll}}$), and pitch ($M_{\rm{pitch}}$) moments, are calculated by
\begin{equation}
\begin{bmatrix}
    F_{z} \\ M_{\rm{roll}} \\ M_{\rm{pitch}}
\end{bmatrix}
=
\begin{bmatrix}
    1&1\\d_{\rm{roll}}&d_{\rm{roll}}\\d_{\rm{pitch}}&d_{\rm{pitch}}
\end{bmatrix}
   \begin{bmatrix}
       L(\theta_{l}, V) \\  L(\theta_{r}, V)
   \end{bmatrix}
\end{equation}
with $d_{\rm{pitch}}$ and $d_{\rm{roll}}$ as the respective moment arms.
which are controllable functions of $\theta$, with $V$ the swimming speed. The function, $L(\theta,V)$, is non-linear because of the $c_{L}({\theta})$ and $V^{2}$ terms, which complicates the control system. The first term is linear for small angles of attack up to stall. Nevertheless, it is a strictly increasing function up to stall and has a polynomial approximation. The control authority of the pitch and roll is guaranteed with sufficiently large forward swimming speed. 
\subsection{Actuation of Yaw}
\begin{figure}[t]
    \centering
    \includegraphics[ width  = \linewidth]{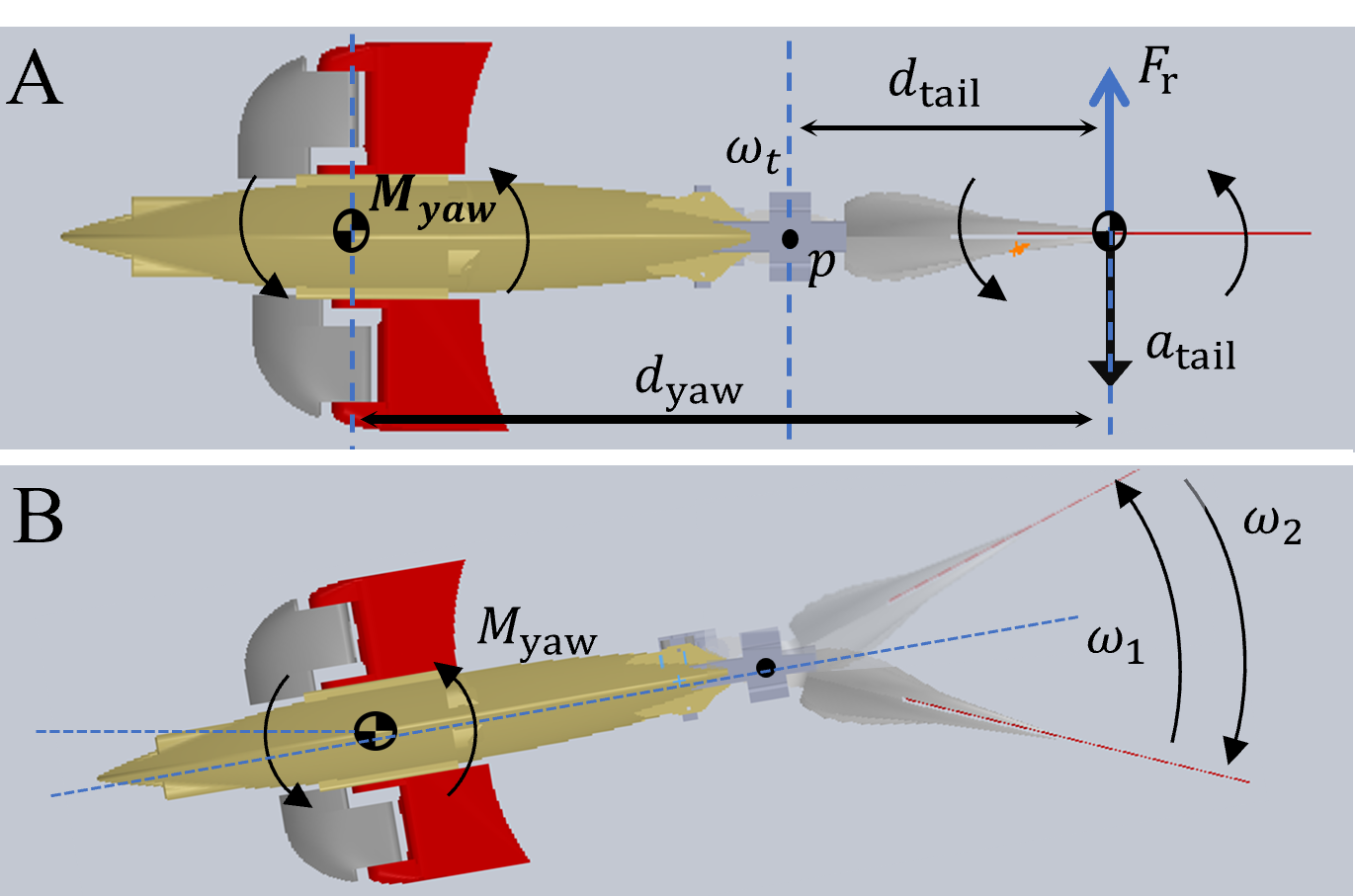}
    \caption{Yaw forces on the fish. (A) Active torque applied by the caudal fin on the water. (B) Left turn from a net clockwise $M_{\rm{yaw}}$ with $\omega_{\rm{2}} > \omega_{\rm{1}}$. }
    \label{fig:yaw moment}
\end{figure}
The yaw actuation of Snapp is an underactuated mechanism as the forward thrust and the yaw-turn moment are accomplished by the caudal fin. Ng et. al \cite{caudal_turning} attempted to course correct the lateral swimming drift on their scotch-yoke mechanism using a speed differential while \cite{SoroFastEscape} demonstrated the key to fast turns is a rapid first stroke in a C shape (phase II-III) followed by a slower return stroke (phase IV-I). We built on these ideas and proposed a cyclic differential algorithm that can have fast turns while maintaining swimming speed. We did this by modulating the motor voltage,
\begin{equation}
    u_t =\left( u_a + u_s \rm{sign}(\sin(\theta + \phi + \gamma_0)) \right)V_{\rm{t}}
    \label{equation:diff}
\end{equation}
where $u_t$ is the total voltage applied on the motor; $u_a$ is the average voltage to produce thrust  and is proportional to the steady-state motor speed $\omega_{\rm{m}}$; $u_s$ is the differential amplitude of the square-wave voltage controlled by the user; $\theta_{\rm{m}}$ is the angular position of motor measured by an encoder which indicates the position of caudal-fin; $\phi$ is the initial phase of the sin function; $\gamma_0$ is a constant angle to compensate the response delay and can be calibrated; and $V_{\rm{t}}$ the terminal voltage. The square-wave voltage is then generated by applying a sign function on a sine function.

This results in a fin stroke with different speeds. To achieve a left turn (depicted in Fig. \ref{fig:yaw moment}(B)), the first stroke is slower with the tail angular speed, $\omega_{\rm{1}}$, and the return stroke with a faster tail angular speed, $\omega_{\rm{2}}$, and vice versa for the right turn. Fig. \ref{fig:yaw moment} shows the respective motions and their forces.

 To find $M_{\rm{yaw}}$, we need to find the water reaction forces, $F_{\rm{r}}$, acting with a moment arm, $d_{\rm{yaw}}$, on the tail fin. With the added mass principle\cite{Virtual_mass}, we get the following
 \begin{equation}
     \begin{aligned}
        F_{r} &= m_{w} a_{\rm{tail}}\\
        M_{\rm{yaw}} &=  d_{\rm{yaw}} F_{r}
     \end{aligned}
    \label{equation:Myaw}
\end{equation}
where $m_{\rm{w}}$, the virtual mass accelerated with the same acceleration as the tail, $a_{\rm{tail}}$, which is proportional to $\omega^{\rm{2}}_{\rm{m}}$ (\ref{equation: tail acceleration}). The yaw moment per tail beat, $M_{\rm{cycle}}$, is derived by taking the difference over the phases as in Fig. \ref{fig:actuation and phases}, with (phase II-III) and right stroke (phase IV-1) as implemented by (\ref{equation:diff}), with the corresponding angular speeds $\omega_{\rm{1}}$ and $\omega_{\rm{2}}$ respectively. We finally derive $M_{\rm{cycle}}$ as a function of $u_{\rm{s}}$
\begin{equation}
    \begin{aligned}
        M_{\rm{cycle}} &= M(\omega_{\rm{1}}) - M(\omega_{\rm{2}}) = M(u_s)
    \end{aligned}
\end{equation}
By modulating the difference in motor speed between tail beat cycles, we can control the overall turn rate per stroke. In the absence of the differential, the swimming gait is symmetric, i.e. the lateral swings of the phases are equal in magnitude and cancel out.
\section{Results and Discussion}
To achieve 3D maneuverability, responsive attitude control is required. To demonstrate this, we conducted experiments for each specific attitude, pitch, roll, and yaw. The angular positions and velocities are sampled at 400Hz, and video recordings above and underwater were taken to track the trajectory of the robot. The results highlight Snapp's system response and its trajectory.

Snapp was tested at two tail actuation at two modes, 2.7 Hz and 4 Hz, determined by measuring the time to complete a 25 m swim course. The average swimming speed associated with the 2.7 Hz and 4Hz were  1m/s and 1.5 m/s respectively, demonstrating a 50\% increase in swimming speed.
For each attitude, we tested the control inputs (\ref{controlinputs}) with pitch, roll and yaw governed by $\;u_{s}\; u_{r}\; u_{p}$ respectively, with the swimming speed kept constant. 
The roll and pitch tests were conducted at a 1 m/s speed. 
This establishes the minimum speed required for effective control. 
Finally, we swam Snap in a figure-8 pattern with an initial steady-state pitch angle of 60$^{\circ}$ to test the robustness of the control methods to disturbances.

\subsection{Field Experiments and Protocols}
The experiments were carried out in a swimming pool of 25 m $\times$ 25 m with an average depth of 1.2 m. Two cameras captured the video footage of the swimming fish. The first camera was positioned at the starting point of the test, and the second underwater camera was positioned parallel to the swimming direction. A drone camera captured overhead footage for a figure-8 swim. The pilot controls the fish in all experiments.

\subsection{Pitch and Dive Control}
\begin{figure}[t]
    \centering
    \includegraphics[width=\linewidth]{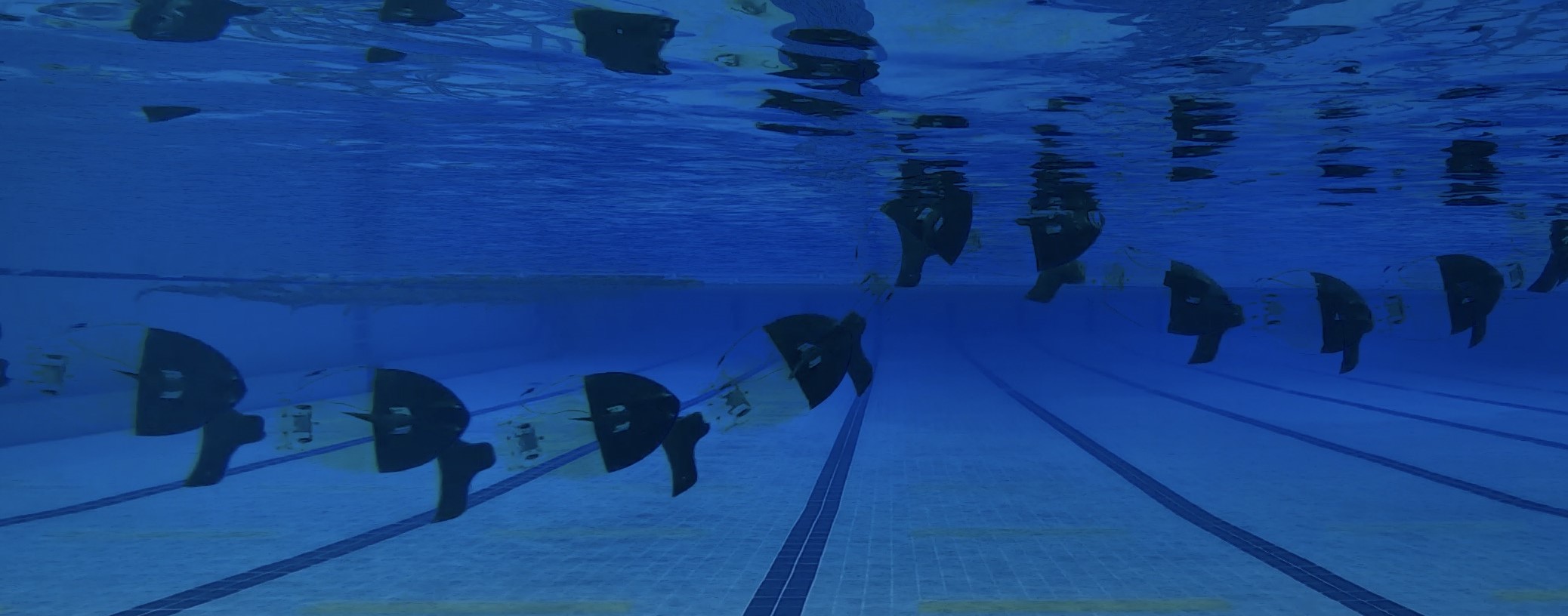}
    \caption{An overlay of the dive experiment from the stationary underwater camera. The image captures the dive and lift sequence of the robot fish. }
    \label{fig:divepic}
\end{figure}

The emphasis of this experiment is to control the dive and ascend actions of Snapp. The pectoral fins exert a lift force and a moment as mentioned in section III(A). These forces contribute to the vertical translation and pitch rotation. To showcase this, Snapp performed the U-shape trajectories in Fig. \ref{fig:divepic}. The control commands controlled the pitching moments. 
\begin{figure}[t]
    \centering
    \includegraphics[width=\linewidth]{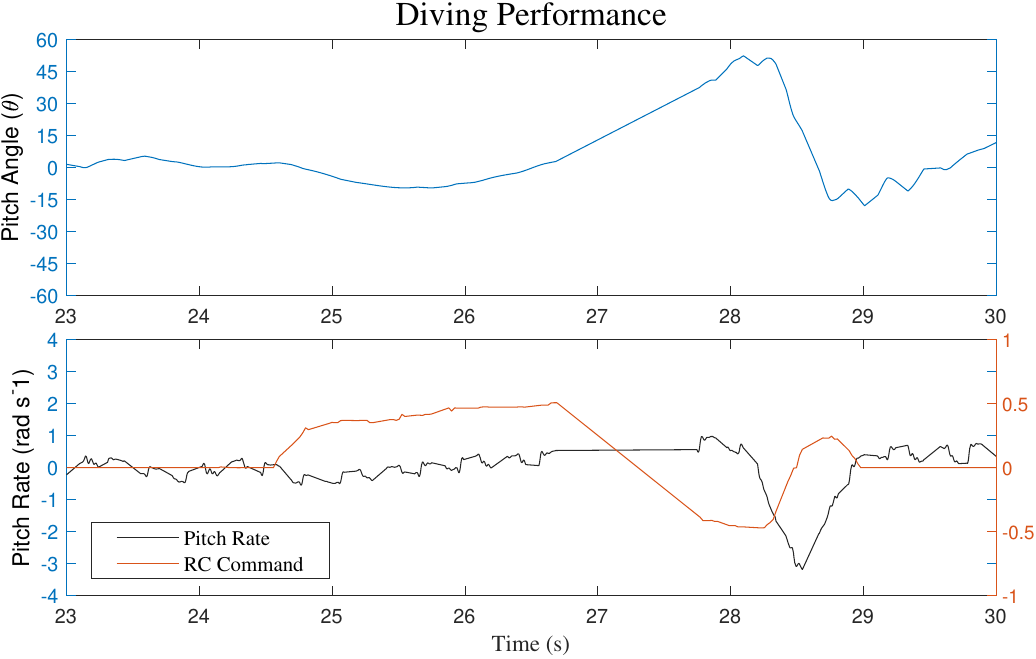}
    \caption{ A plot of the angular position and angular rate with respect to time given the user commands. The pitch is set in an inverted Y mode. When the pitch stick is lowered (-) the pectoral fins are lowered, creating an upward lift, increasing the pitch angle, and providing vertical lift. Angular pitch rate (bottom) responds timely to the dive command, and the resurface command. The RC commands are configured in an inverse Y.}
    \label{fig:divedata}
\end{figure}
Snapp initially swam without commands, followed by a dive, then an ascend command. Fig. \ref{fig:divedata} highlights the relationship between the pitch command, and its angle and rate. At the 24.5 s mark, a positive pitch command (dive) of 0.5 was issued. Snapp descended with a maximum pitch angle of -10$^{\circ}$ and then leveled back to 0$^{\circ}$. This dive action corresponded to a maximum positive pitch rate of 0.5 rad/s as seen in the rate plot of Fig. \ref{fig:divedata}. Snapp then ascended with a maximum rate of -3 rad/s (average of 1.5 rad/s) following the ascend command. Finally, at the 28.5 s mark, there was a rapid increase in pitch rate following the ascending stage. This rapid increase is resulted from Snapp leaping out of the water and later falling due to gravity. 
 \begin{figure*}[ht]
    \centering
    \includegraphics[width = 7 in]{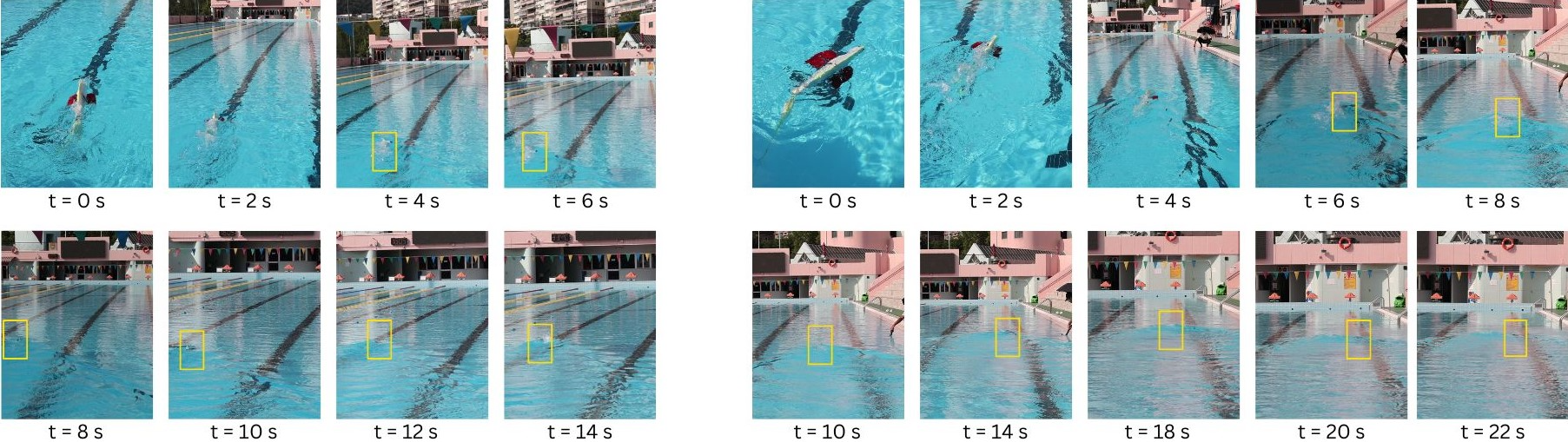}
    \caption{(Left) Snapp swimming a 25 m distance without any compensation. The robot fish drifts off to the left side.  (Right) Snapp swimming 25 m with roll compensation. Snapp is capable of swimming within the boundary of the 2.5 m lanes. 
     \label{fig:roll image}
   }
\end{figure*}
\begin{figure}[ht]
    \centering
    \includegraphics[width = \linewidth ,height = 2.4in]{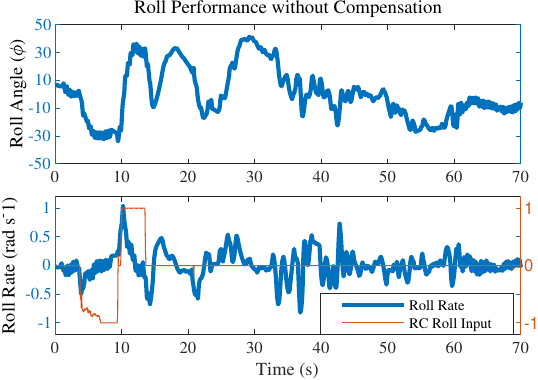}
    \caption{ Step Response and natural drift: An initial short impulse was applied to test the response of the system. The roll changes are immediate and responsive. Without any active input, there is a large natural drift to the left caused by the roll instability.   }
    \label{fig:rollnocomp}
\end{figure}
\begin{figure}[ht]
    \centering
    \includegraphics[width=\linewidth,height = 2.4in]{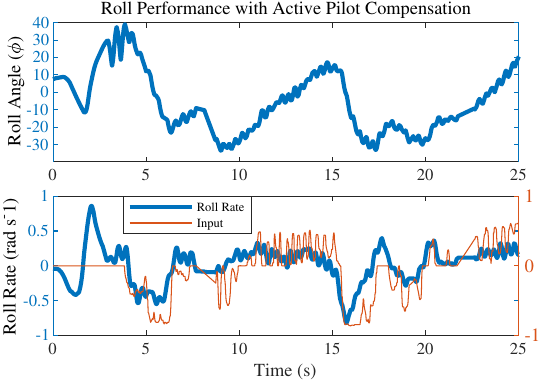}
    \caption{Active roll compensation controlled by the pilot. By controlling the roll, the pilot is capable of controlling the yaw. A positive roll induces a positive yaw. Data showcases the commands throughout the swim and their corresponding roll angle, and roll rates. }
    \label{fig:rollwithcomp}
\end{figure}
Despite the commands between dive and ascend having the same magnitudes (0.5), they have different pitch rates. There are two major reasons for this: (i) an initial offset pitch and (ii) a pitching moment from the caudal fin. Both factors contributed to a positive pitch of Snapp. The former was caused due to water seeping into the shell, thus having a moment imbalance, and the latter is caused by the thrust vector acting below the center of mass, inducing an upward pitch moment. Nevertheless, the pilot completed the U-shape trajectory despite the disturbances.
\subsection{Roll Control and Straight Line Swimming}

The roll test was conducted by measuring the roll angles and rates over a 25 m swim course. It is observed that a change in roll causes Snapp to deviate from a straight line. This is because Snapp has a net pitching moment from the caudal fin as the thrust force vector acts below the center of mass. The combined pitching and rolling moments induced a lateral drift. Fig. \ref{fig:roll image}(left) highlights the natural lateral drift that occurs without compensation. 

To test the pectoral fin's ability to control roll, we tested (i) a step response, (ii) a straight swim without compensation, and (iii) a straight swim with roll compensation. The roll angle and rate are measured with the user commands. The commands control the roll moments, which reach equilibrium due to the buoyancy moment. 

The step response of the roll command was observed in Fig. \ref{fig:rollnocomp}. The results showed a 35$^{\circ}$ angle in their respective directions in 2s. The first step input has a maximum roll rate of -0.5 rad/s, and the second has a maximum roll rate of 1 rad/s. Snapp then swam without compensation. The roll drifted and eventually swam out of bounds of the test site with a lateral drift of 5m from the starting point. 
In contrast, Fig. \ref{fig:rollwithcomp} shows the step response when active roll compensation is applied by the pilot, highlighting the pilot's attempt to stabilize the roll angle.
The roll compensation has a response time of 0.5 seconds and a maximum angular rate of 0.6 rad/s at a swimming speed of 1 m/s. 
Note that the roll rates followed the commands closely as the pilot adopted a bang-bang (10 - 15 s) control method. AS a result, Snapp swam within the 2.5 m bounds of a standard swimming lane.
\subsection{Yaw Control for Straight Line Swim }
\begin{figure*}[t]
    \centering
    \includegraphics[width=\linewidth]{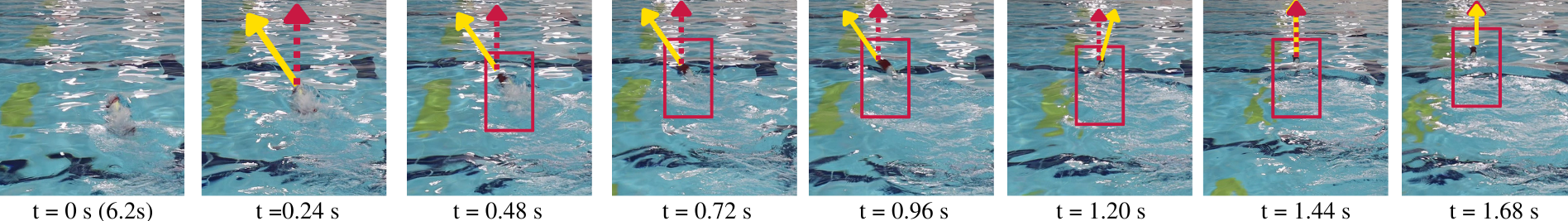}
    \caption{Snapp performing a sharp yaw correction. Snapp was initially headed north-west (yellow line) with the red line (dotted) as reference north. Within a single tail-beat (0.96 - 1.44 s), it corrects its heading to complete the full 25m swim. Note that the left yellow marker of the swimming pool is the same in all pictures. }
    \label{Fig:Sharpyaw pic}
\end{figure*}
\begin{figure}[t]
    \centering
    \includegraphics[ width = \linewidth]{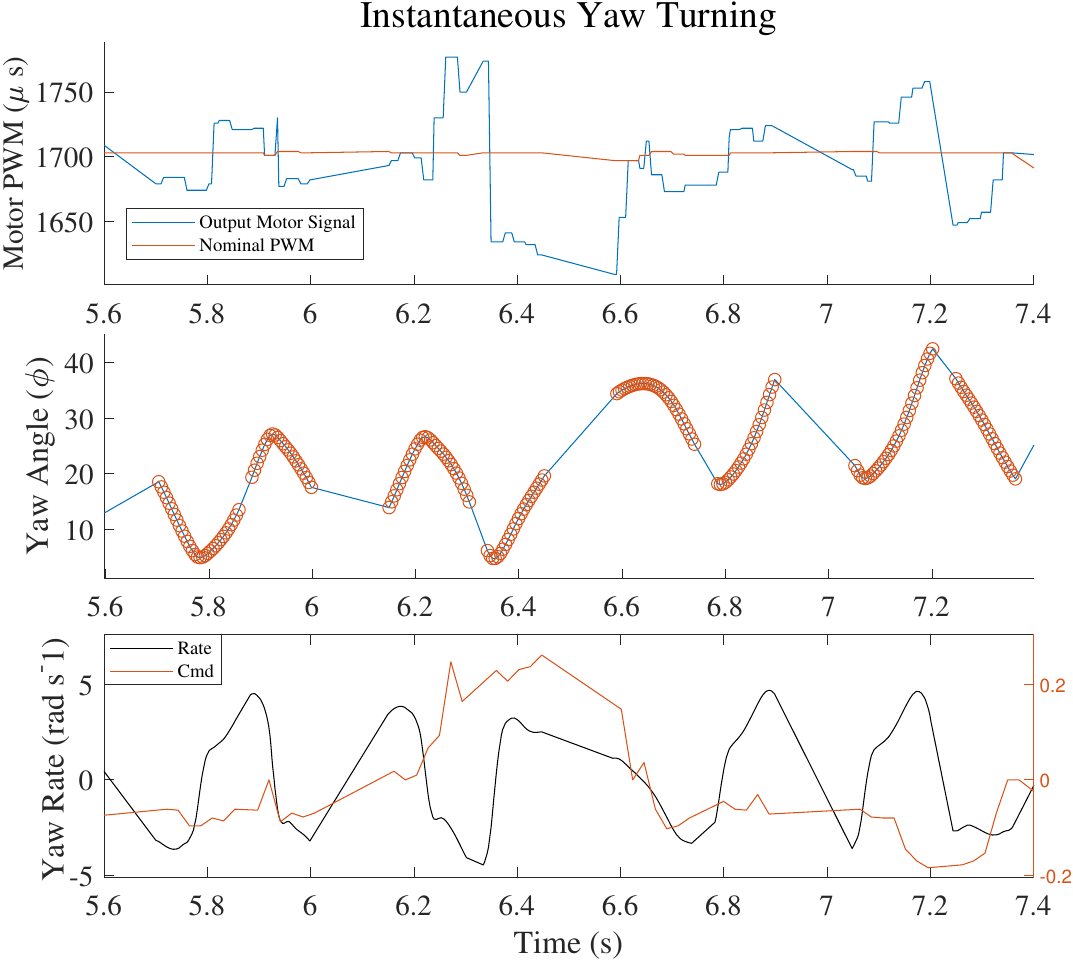}
    \caption{A sharp yaw correction. (Top) Differential algorithm in action represented by motor output. (Middle) The yaw angle of the swimming fish. (Bottom) The corresponding yaw rates to the instantaneous correction. The command was executed at the 6.2 s - 6.6 s mark.}
    \label{fig:fastyaw}
\end{figure}
\begin{figure}[t]
    \centering
    \includegraphics[ width = \linewidth]{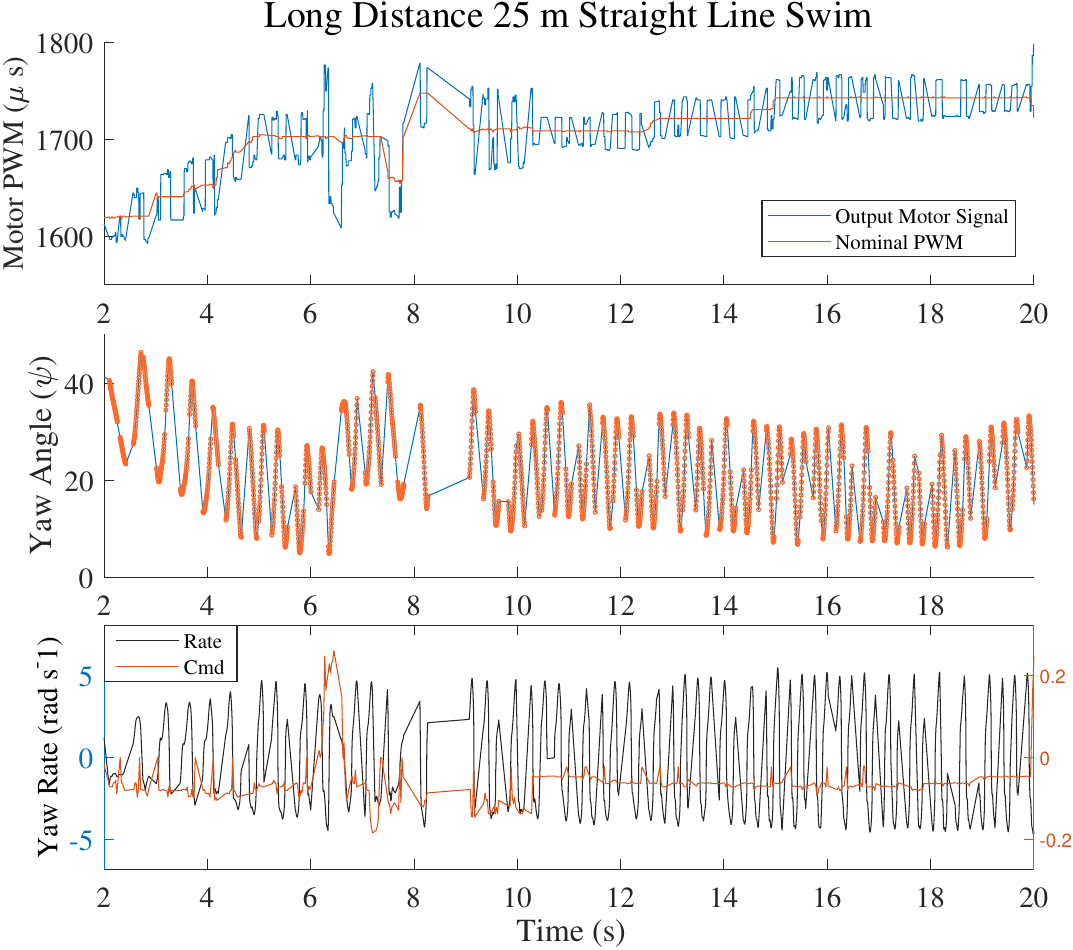}
    \caption{Full experiment over 25 m. (Top) Motor signals and the modulated signals from the cyclic-differential method. (Middle) Yaw angles of the entire swim. (Bottom) Yaw rates and their corresponding commands.}
    \label{Fig: yaw straight swim}
\end{figure}
\begin{figure}[ht!]
    \centering
    \includegraphics[width=1\columnwidth]{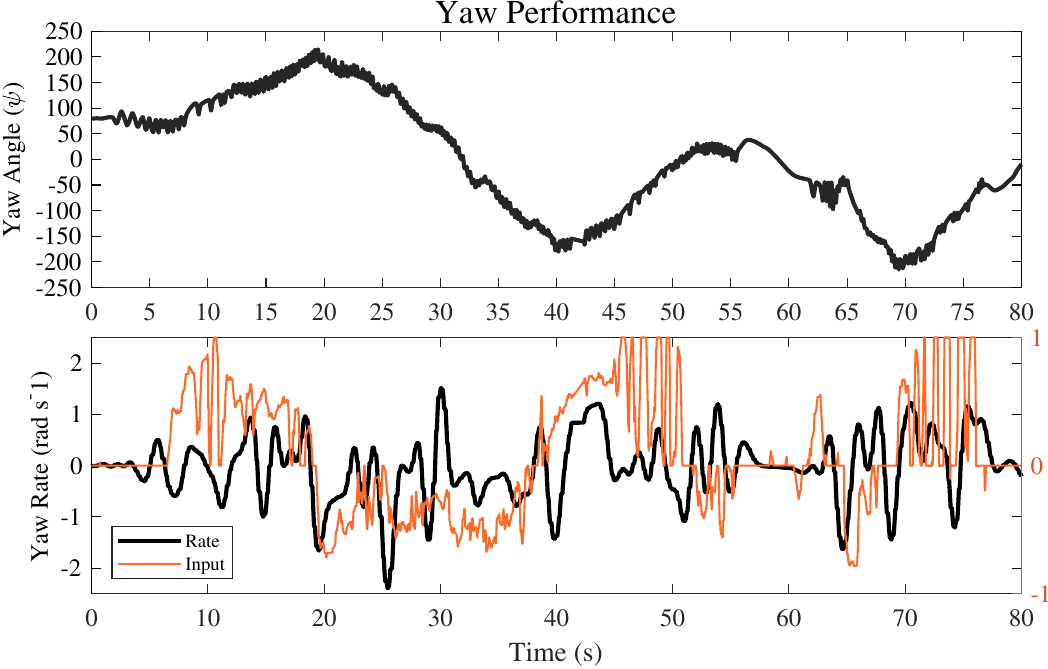}
    \caption{figure-8 and S curve data set. (Top) The yaw angle and the pilot's input. (Bottom) The yaw rate closely follows the command by the pilot. The yaw rate was averaged over 2 tail beats. The figure-8 and S-curve were done in one take.}
    \label{fig:figure8andScurve}
\end{figure}
\begin{figure}[ht!]
    \centering
    \includegraphics[width=1\columnwidth]{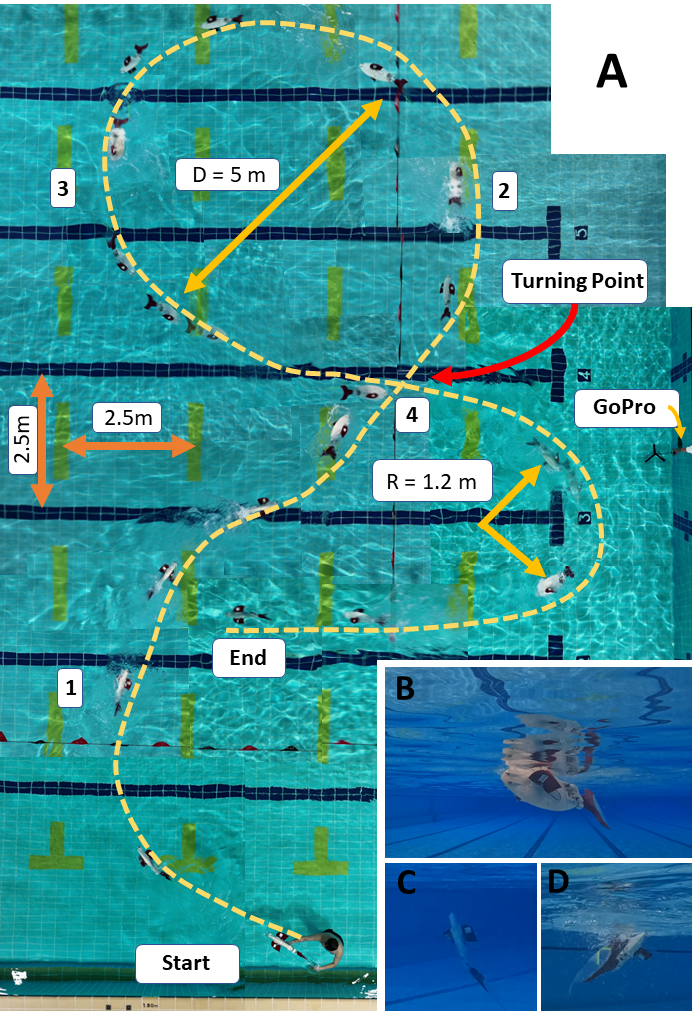}
    \caption{(A) Aerial view of the figure-8 experiment. The fish swims from the start point, follows the order 1-2-3-4, and finally to the endpoint. The image was generated by manually piecing together different snapshots of the video captured by an aerial drone. (B) Snapp making a turn using a combined roll and yaw. (C) Snapp's neutral state with negative buoyancy and a 90$^{\circ}$ pitch angle. (D) Snapp exits circle 4 and accelerates out. The caudal fin is flexed for maximum propulsion.}
    \label{fig:figure8 traj}
\end{figure}
\begin{figure}[ht!]
    \centering
    \includegraphics[width=1\columnwidth]{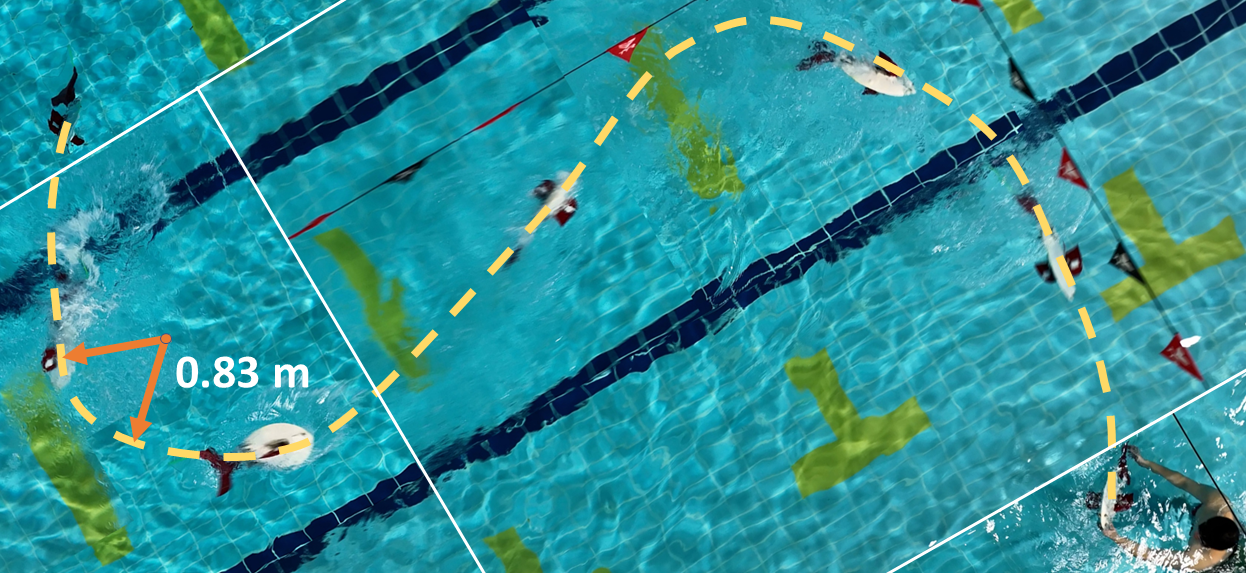}
    \caption{Aerial view of the S-curve experiment. Snapp swims from the end of figure-8 and completes the S-curve to return to the starting point of the experiment. }
    \label{fig:S-curve}
\end{figure}
We tested the cyclic-differential method (\ref{equation:diff}) for instantaneous and cruising turns. The pilot was tasked to swim northwest, followed by a sharp correction to north, and then to swim to the endpoint at maximum speed. Fig. \ref{Fig:Sharpyaw pic} and \ref{fig:fastyaw} highlight a sharp yaw turn executed within one tail beat. This command is executed at the 6.2 - 6.6 s of the swim with a peak-to-peak motor speed differential of 6.24 rad/s (25\% of maximum motor speed at 4 Hz). Note that the maximum differential corresponded with a peak-to-peak voltage difference of 20V or an 80 \% speed difference.

This command, $u_s$, with value 0.2, changed the angle from an 15$^{\circ}$ (avg) to a 28$^{\circ}$(avg) in 0.4 s, with an yaw rate(avg) of 40$^{\circ}$/s. Coinciding with the commands, the angular rate in Fig. \ref{fig:fastyaw}(bottom) had a short negative section (6.2 s - 6.4 s), followed by a much longer positive yaw rate (6.4 s - 6.6 s) before returning to the original swimming pattern.
Fig. \ref{Fig: yaw straight swim} shows the full timeline of the straight swim. Snapp responded to a left turn command (-0.08) at the (2-6 s) mark. 
The yaw angle changed from 40$^{\circ}$ to 15$^{\circ}$ in that period, with a yaw rate (avg) of 6.25 $^{\circ}$/s. 
The latter part showed the pilot's small adjustments to complete the straight swim with 25$^{\circ}$ as the target heading. 
This shows effectiveness of our cyclic-differential method in controlling yaw.

\subsection{Open Water Swimming: Figure-8 and S curve.}
Snapp completed a figure-8 trajectory followed by an S-curve using all control commands simultaneously. 
A person piloted the experiment. 
The commands with their respective angles and velocities were recorded and plotted in Fig. \ref{fig:figure8andScurve}. 
Fig. \ref{fig:figure8 traj}(A) and Fig. \ref{fig:S-curve} depicts the figure-8 attempt and the S-curve respectively.

Initially, Snapp was held by a person and then released when the tail begins to move. Its natural state was to sink and to have a 90$^{\circ}$ pitch angle as in Fig. \ref{fig:figure8 traj}(C). 
This state approximates the uncertainty of open sea swimming, where the initial balance of the fish could be disrupted. 
Despite the disturbances, the pilot completes the figure-8 and S-curve trajectory.

Fig. \ref{fig:figure8andScurve} shows the data collected in this maneuver. 
The 20 s corresponds to the first half of the circle, and the next 20 s corresponds to a full circle, from 200$^{\circ}$ to -180$^\circ$. 
The 60-80 s corresponds to the S-curve as the fish returned to the starting point. 
The 20s mark is the first turning point of the figure-8. 
At this point, the turn rate changed from 1 rad/s to -1.5 rad/s, following the command closely. 
Similar trends occured throughout the swim with the 40s mark representing the second turning point.
The average yaw is calculated from the gradient of the yaw-angle graph. At the 60 s mark, the pilot initiates a burst swim followed by a turn maneuver (65 s) and resulted in a sharp turn of 95$^{\circ}$ in 2 s, a turn rate of 46.7 $^{\circ}$/s (0.81 rad/s).
The top loop in Fig. \ref{fig:figure8 traj} has a diameter of 5 m, and the bottom loop has a turn radius of 1.2 m. The minimum turn radius of 0.83 m (1 BL/s) is observed in the S-curve. Fig. \ref{fig:figure8 traj}(B, C, D) shows the different pitches and rolls during this swim as controlled by the pilot.
Finally, we successfully swam Snapp in the open sea of Hong Kong's Repulse Bay beach. These experiments are available in the video link.
\section{Limitations and Future Works}
The high speeds of Snapp were made possible due to the large motor of 150 W. While it achieves a high speed at this mass of 5 kg, it has a low BL/s despite the high power input suggesting a low efficiency. 
Future improvements to the design may include reduction of weight, optimization of swimming gait patterns, and exploration of different tail designs to increase the overall swimming efficiency for prolonged mission times.

Nonetheless, Snapp's remarkable advancements in speed and maneuverability have pushed the boundaries of robotic fish performance in open-water environments. Future work should focus on developing feedback systems for autonomous swimming, thus enabling and enhancing various applications such as observing coral reefs, aquatic life, underwater search-and-rescue missions, monitoring water quality, assisting in marine infrastructure maintenance, and performing stealthy surveillance in maritime security.

\bibliographystyle{./bibliography/IEEEtran}
\bibliography{./bibliography/IEEEabrv,./bibliography/ref}

\begin{thebibliography}{10}
\providecommand{\url}[1]{#1}
\csname url@samestyle\endcsname
\providecommand{\newblock}{\relax}
\providecommand{\bibinfo}[2]{#2}
\providecommand{\BIBentrySTDinterwordspacing}{\spaceskip=0pt\relax}
\providecommand{\BIBentryALTinterwordstretchfactor}{4}
\providecommand{\BIBentryALTinterwordspacing}{\spaceskip=\fontdimen2\font plus
\BIBentryALTinterwordstretchfactor\fontdimen3\font minus
  \fontdimen4\font\relax}
\providecommand{\BIBforeignlanguage}[2]{{%
\expandafter\ifx\csname l@#1\endcsname\relax
\typeout{** WARNING: IEEEtran.bst: No hyphenation pattern has been}%
\typeout{** loaded for the language `#1'. Using the pattern for}%
\typeout{** the default language instead.}%
\else
\language=\csname l@#1\endcsname
\fi
#2}}
\providecommand{\BIBdecl}{\relax}
\BIBdecl

\bibitem{LauderSummary}
G.~V. Lauder, E.~J. Anderson, J.~Tangorra, and P.~G.~A. Madden, ``Fish
  biorobotics: kinematics and hydrodynamics of self-propulsion,'' \emph{J. Exp.
  Biol.}, vol. 210, no.~16, pp. 2767--2780, 08 2007.

\bibitem{swimspeedtuna}
B.~A. Block, D.~Booth, and F.~G. Carey, ``{Direct Measurement of Swimming
  Speeds and Depth of Blue Marlin},'' \emph{J. Exp. Biol.}, vol. 166, no.~1,
  pp. 267--284, 05 1992.

\bibitem{Locomotion}
T.~Castro-Santos, E.~Goerig, P.~He, and G.~V. Lauder, ``Applied aspects of
  locomotion and biomechanics,'' ser. Fish Physiology.\hskip 1em plus 0.5em
  minus 0.4em\relax Academic Press, 2022, vol.~39, pp. 91--140.

\bibitem{weRofi}
X.~Liao, C.~Zhou, Q.~Zou, J.~Wang, and B.~Lu, ``Dynamic modeling and
  performance analysis for a wire-driven elastic robotic fish,'' \emph{IEEE
  Robot. and Automat. Lett.}, vol.~7, no.~4, pp. 11\,174--11\,181, Oct 2022.

\bibitem{TensengrityFish}
B.~Chen and H.~Jiang, ``Swimming performance of a tensegrity robotic fish,''
  \emph{Soft Robotics}, vol.~6, no.~4, pp. 520--531, 2019.

\bibitem{fishturn_IJAR}
Z.~Li, L.~Ge, W.~Xu, and Y.~Du, ``Turning characteristics of biomimetic robotic
  fish driven by two degrees of freedom of pectoral fins and flexible
  body/caudal fin,'' \emph{Int. J. of Adv. Robot. Syst.}, vol.~15, no.~1, 2018.

\bibitem{Sofi}
R.~K. Katzschmann, J.~DelPreto, R.~MacCurdy, and D.~Rus, ``Exploration of
  underwater life with an acoustically controlled soft robotic fish,''
  \emph{Sci. Robot.}, vol.~3, no.~16, p. eaar3449, 2018.

\bibitem{LeapingFish}
D.~Chen, Z.~Wu, Y.~Meng, M.~Tan, and J.~Yu, ``Development of a high-speed
  swimming robot with the capability of fish-like leaping,'' \emph{IEEE/ASME
  Trans. on Mechatronics}, vol.~27, no.~5, pp. 3579--3589, 2022.

\bibitem{tunabot}
J.~Zhu, C.~White, D.~K. Wainwright, V.~D. Santo, G.~V. Lauder, and
  H.~Bart-Smith, ``Tuna robotics: A high-frequency experimental platform
  exploring the performance space of swimming fishes,'' \emph{Sci. Robot.},
  vol.~4, no.~34, p. eaax4615, 2019.

\bibitem{OpenFish}
S.~C. {van den Berg}, R.~B. Scharff, Z.~Rusák, and J.~Wu, ``Openfish:
  Biomimetic design of a soft robotic fish for high speed locomotion,''
  \emph{HardwareX}, vol.~12, p. e00320, 2022.

\bibitem{recoil}
P.~W. Webb, ``{Is the High Cost of Body/Caudal Fin Undulatory Swimming due to
  increased Friction Drag or Inertial Recoil?}'' \emph{J. Exp. Biol.}, vol.
  162, no.~1, pp. 157--166, 01 1992.

\bibitem{swim_stability}
P.~W. Webb and D.~Weihs, ``Stability versus maneuvering: Challenges for
  stability during swimming by fishes,'' \emph{Integrative and Comparative
  Biology}, vol.~55, no.~4, pp. 753--764, 2015.

\bibitem{CasiTuna}
S.~Du, Z.~Wu, J.~Wang, S.~Qi, and J.~Yu, ``Design and control of a
  two-motor-actuated tuna-inspired robot system,'' \emph{IEEE Trans. Syst. Man
  Cybern.}, vol.~51, no.~8, pp. 4670--4680, Aug 2021.

\bibitem{cuhk}
Y.~Zhong, Z.~Li, and R.~Du, ``A novel robot fish with wire-driven active body
  and compliant tail,'' \emph{IEEE/ASME Trans. on Mechatronics}, vol.~22,
  no.~4, pp. 1633--1643, 2017.

\bibitem{MagnetFish}
D.~Romano, A.~Wahi, M.~Miraglia, and C.~Stefanini, ``Development of a novel
  underactuated robotic fish with magnetic transmission system,''
  \emph{Machines}, vol. 10(9), p. 755, 2022.

\bibitem{lighthill_1970}
M.~J. Lighthill, ``Aquatic animal propulsion of high hydromechanical
  efficiency,'' \emph{J. Fluid Mech.}, vol.~44, no.~2, p. 265–301, 1970.

\bibitem{Virtual_mass}
L.~M. J., ``Hydromechanics of aquatic animal propulsion,'' \emph{Annu. Rev. of
  Fluid Mech.}, vol.~1, no.~1, pp. 413--446, 1969.

\bibitem{SoroFastEscape}
A.~D. Marchese, C.~D. Onal, and D.~Rus, ``Autonomous soft robotic fish capable
  of escape maneuvers using fluidic elastomer actuators,'' \emph{Soft
  Robotics}, vol.~1, no.~1, pp. 75--87, 2014, pMID: 27625912.

\bibitem{DragReduction}
D.~S. Barett, M.~S. Triantafyllou, D.~K.~P. YUE, M.~A. Grosenbaugh, and M.~J.
  Wolfgang, ``Drag reduction in fish-like locomotion,'' \emph{J. Fluid Mech.},
  vol. 392, p. 183–212, 1999.

\bibitem{worldrecordfish}
W.~Z. Gupta S. Ng JKT Shen~Z, ``Development of high performance mechanical
  robotic fish,'' 2018, adv. Maritime Eng. Conf. ).

\bibitem{Strouhal}
R.~.~T. Taylor~G., Nudds, ``Flying and swimming animals cruise at a strouhal
  number tuned for high power efficiency,'' \emph{Nature}, vol. 425, pp.
  707--711, 10 2003.

\bibitem{caudal_turning}
W.~Z. Ng~JKT, Tang~CC, ``Caudal-fin based turning mechanism for an
  underactuated fast swimming robotic fish,'' 2021, adv. Maritime Eng. Conf.
  SPB,Rus.

\end{thebibliography}
\end{document}